\documentclass[conference, a4paper]{IEEEtran}
\IEEEoverridecommandlockouts
\usepackage[top=0.70in,bottom=1in,left=0.62in,right=0.62in]{geometry}
\usepackage{cite}
\usepackage{amsmath,amssymb,amsfonts}
\usepackage{graphicx}
\usepackage{subfigure}
\usepackage{textcomp}
\usepackage{xcolor}
\usepackage{eso-pic}
\usepackage{titlesec}
\usepackage{url}
\usepackage{multicol}
\usepackage{multirow}
\usepackage{amsmath}
\usepackage{booktabs}
\usepackage{algpseudocode}
\usepackage{lipsum}
\usepackage{fancybox}
\usepackage{listings}
\usepackage[utf8]{inputenc}
\usepackage{comment}
\usepackage{hyperref}
\usepackage[ruled,vlined]{algorithm2e}
\SetKwComment{Comment}{$\triangleright$\ }{}

\titleformat{\subsection}{\normalsize\bfseries}{\thesubsection}{1em}{}

\def\BibTeX{{\rm B\kern-.05em{\sc i\kern-.025em b}\kern-.08em
    T\kern-.1667em\lower.7ex\hbox{E}\kern-.125emX}}

\def\thesection{\arabic{section}}  
\def\thesubsection{\thesection.\arabic{subsection}}
\def\thesubsubsection{\thesubsection.\arabic{subsubsection}}

\def\thesectiondis{\thesection}                     
\def\thesubsectiondis{\thesectiondis.\arabic{subsection}}
\def\thesubsubsectiondis{\thesubsectiondis.\arabic{subsubsection}}

\makeatletter

\renewcommand\section{\@startsection {section}{1}{\z@}%
    {-3.5ex \@plus -1ex \@minus -.2ex}
    {2.3ex \@plus.2ex}
    {\normalfont\normalsize\bfseries\raggedright}} 

\def\@makechapterhead#1{%
    \vspace*{50\p@}
    {\parindent \z@ \raggedright \normalfont
     \huge\bfseries \thechapter\hspace{1em}#1\par\nobreak
     \vskip 40\p@}}

\def\ps@IEEEtitlepagestyle{%
    \def\@oddfoot{\mycopyrightnotice}%
    \def\@evenfoot{}%
}

\def\mycopyrightnotice{
  {\footnotesize ~\copyright2025 ~IEEE. This paper has been accepted and presented at the IEEE ECAI 2025. The final version will be available in the IEEE Xplore Digital Library.\hfill} 
  \gdef\mycopyrightnotice{}
}

\makeatother
\def\confheader#1{%
    \def\ps@IEEEtitlepagestyle{%
        \old@ps@IEEEtitlepagestyle%
        \def\@oddhead{\strut\hfill#1\hfill\strut}%
        \def\@evenhead{\strut\hfill#1\hfill\strut}%
    }%
    \ps@headings%
}
\makeatother
\begin{document}

\newcommand\AtPageUpperMyright[1]{\AtPageUpperLeft{
 \put(\LenToUnit{0.28\paperwidth},\LenToUnit{-1cm}){
     \parbox{0.78\textwidth}{\raggedleft\fontsize{9}{11}\selectfont #1}}
 }}

\title{Social Media Sentiments Analysis on the July Revolution in Bangladesh: A Hybrid Transformer Based Machine Learning Approach \\
\thanks{\vspace{10pt}\noindent\parbox[b]{\dimexpr\columnwidth-1em}{ 
\textbf{\rule{5.1cm}{1pt}}\\ 
*Corresponding Author: Md. Sabbir Hossen \vspace{2mm} \\  This research is supported by Bangladesh University and NextGen AI Lab; We are grateful for the resource, guidance and encouragement provided by the institution.}}
}
\author{\IEEEauthorblockN{Md. Sabbir Hossen\IEEEauthorrefmark{9}\IEEEauthorrefmark{1}, Md. Saiduzzaman\IEEEauthorrefmark{9}, and Pabon Shaha\IEEEauthorrefmark{6}}
\IEEEauthorblockA{\IEEEauthorrefmark{9}Dept. of Computer Science \& Engineering, Bangladesh University, Dhaka, Bangladesh}
\IEEEauthorblockA{\IEEEauthorrefmark{6}Dept. of Computer Science \& Engineering, Mawlana Bhashani Science \& Technology University, Tangail, Bangladesh}
\IEEEauthorblockA{Email: sabbirhossen5622@gmail.com, saiduzzamancse56bu@gmail.com, and pabonshahacse15@gmail.com} 
}

\maketitle

\begin{abstract}
The July Revolution in Bangladesh marked a significant student-led mass uprising, uniting people across the nation to demand justice, accountability, and systemic reform. Social media platforms played a pivotal role in amplifying public sentiment and shaping discourse during this historic mass uprising. In this study, we present a hybrid transformer-based sentiment analysis framework to decode public opinion expressed in social media comments during and after the revolution. We used a brand new dataset of 4,200 Bangla comments collected from social media. The framework employs advanced transformer-based feature extraction techniques, including BanglaBERT, mBERT, XLM-RoBERTa, and the proposed hybrid XMB-BERT, to capture nuanced patterns in textual data. Principle Component Analysis (PCA) were utilized for dimensionality reduction to enhance computational efficiency. We explored eleven traditional and advanced machine learning classifiers for identifying sentiments. The proposed hybrid XMB-BERT with the voting classifier achieved an exceptional accuracy of 83.7\% and outperform other model classifier combinations. This study underscores the potential of machine learning techniques to analyze social sentiment in low-resource languages like Bangla. 
\end{abstract}

\begin{IEEEkeywords}
Machine Learning, Deep Learning, Sentiment Analysis, Natural Language Processing, Transformer Model \vspace{-2mm}
\end{IEEEkeywords}

\section{INTRODUCTION}
\begin{figure} []
	\centering 
\subfigure{\includegraphics[height=2.28cm, width=4.2cm]{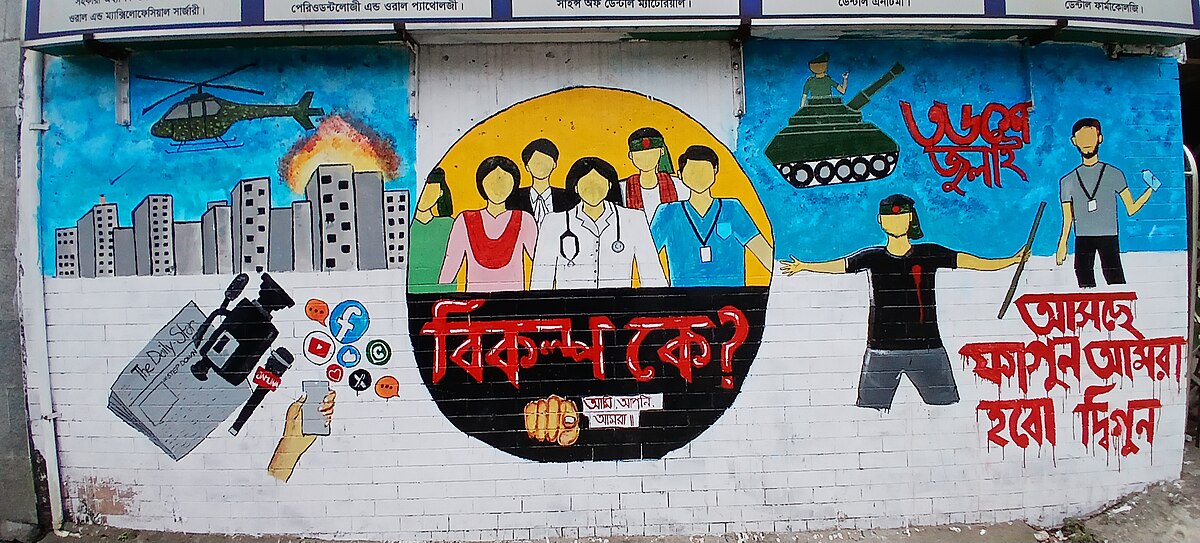}} \hspace{-2mm}
\subfigure{\includegraphics[height=2.28cm, width=4.2cm]{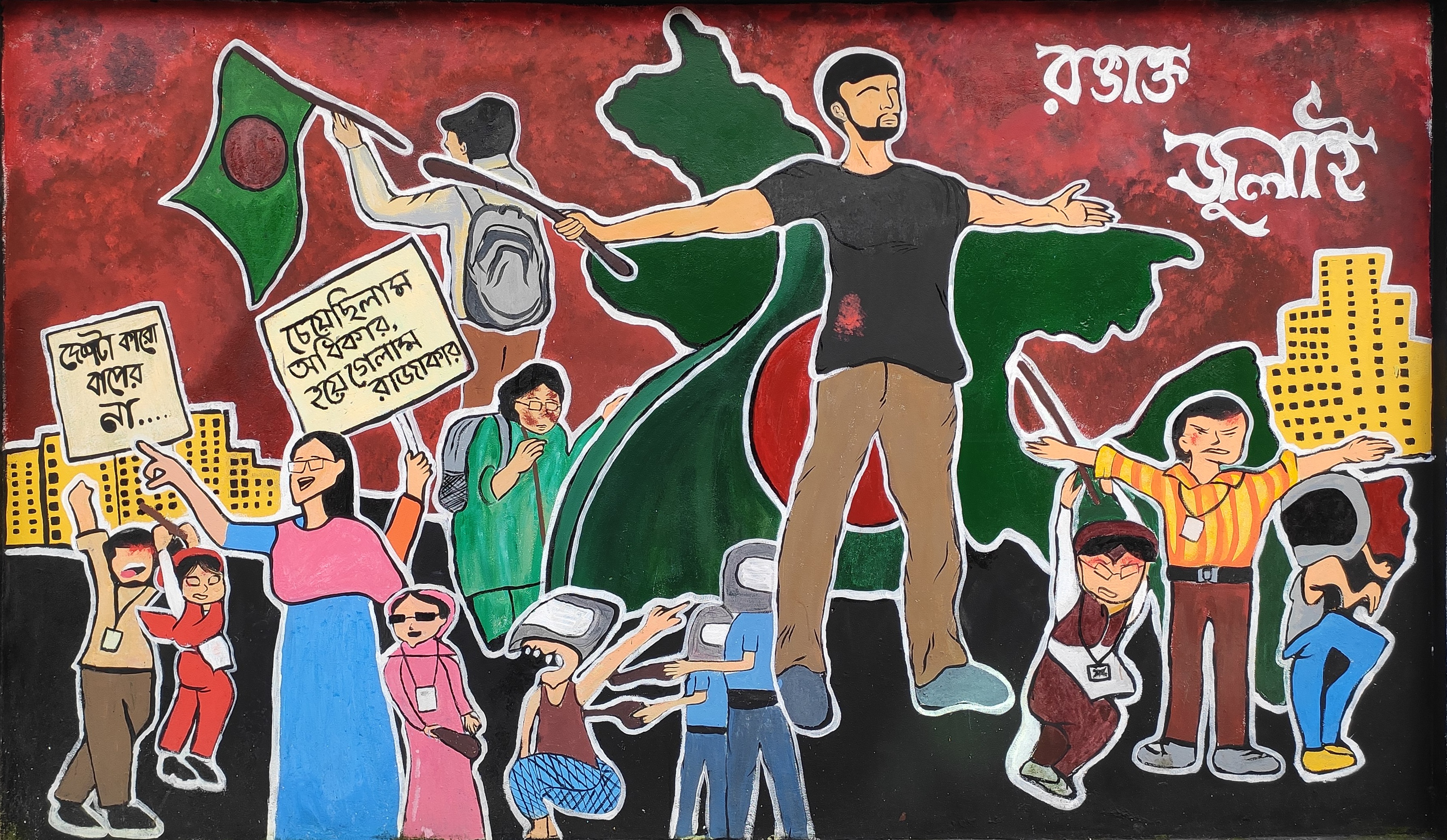}	
} \vspace{-2mm}
		\caption{Graffiti of Revolution and Unity: Visual Narratives of the July Revolution Student Protests in Bangladesh Depicting Resistance, Solidarity, and National Identity Through Artistic Expressions on Public Walls \cite{gft1} \cite{gft2}} \vspace{-5mm}
	\label{fig:gi}
\end{figure}

\vspace{-2mm}The July Revolution in Bangladesh was a historic mass uprising spearheaded by students, supported by people from all walks of life, to demand justice, accountability, and systemic reforms \cite{dstar}. This historic movement, fueled by widespread grievances and social unrest, was a testament to collective resilience and a tragic reminder of the cost of freedom \cite{uprising}. The revolution resulted in a staggering death toll of 1,423 individuals, with an estimated 22,000 others injured. Among them, 587 people were left permanently disabled, while 685 lost partial or complete vision due to gunshot wounds, including 92 individuals who were rendered completely blind in both eyes \cite{july_massacre_2024} \cite{rtv}. The significance of the Revolution is visually captured in publicly available Figure \ref{fig:gi}. Throughout this movement, social media became a powerful tool for rallying support, disseminating information, and documenting events in real time \cite{genz}. Sparked by escalating social and political tensions, the movement gained momentum through widespread protests, rallies, and the strategic use of social media to amplify the voices of the oppressed. Social media played an integral role in shaping public opinion and fostering solidarity throughout the movement. 
The sentiment expressed by users on platforms like Facebook, Twitter (X), and YouTube reflected the anguish and hope of a nation striving for change. Bengali, the fourth most spoken language in the world with over 265 million speakers, remains underrepresented in the field of Natural Language Processing \cite{sen2022bangla}. Despite its widespread use, it is still considered a low-resource language. Sentiment analysis, a method combining natural language processing and machine learning, is instrumental in identifying the emotional tone of textual content and could significantly contribute to understanding public sentiment in Bengali. This study introduced a machine learning framework to analyze public sentiments in Bangla comments on the July Revolution. \textbf{\large{Our major contributions are,}} we collected 4200 comments from public posts in social media during and after the revolution to build the dataset. We utilized several pre-trained transformer-based models for feature extractions, including BanglaBERT, mBERT, and XLM-RoBERTa. Additionally, by combining these three transformer models, we created a hybrid transformer model; we named it Hybrid XMB-BERT. Principle Component Analysis (PCA), t-SNE, and UMAP were utilized for dimensionality reduction and visualization. Traditional and some advanced classifiers were employed to ensure robust performance. Accuracy, precision, recall, and F1 score were used to evaluate the model. The remaining sections of our research are organized as follows. Section II provides a comprehensive review of existing literature on sentiment analysis in Bangla. Section III details the overall building of the proposed model. Section IV presents the experimental setup and Implementation Procedure. Section V presents the critical evaluation of the experimental results, comparing the performance of various machine learning models. Finally, Section VI concludes with a summary of key findings and potential future directions for research. \vspace{-3mm}

\section{LITERATURE REVIEW}
\vspace{-2.8mm}Many researchers have developed models and methods to analyze the sentiment in different ways. However, this field remains underexplored, especially for bangla sentiment analysis task and remains an area of active research and innovation. In this part, Several past studies have been reviewed and summarized to give insight into the current status of this subject in order to fully understand the advancements and challenges. \textit{M.N. Haque et al.} \cite{hoque2024exploring} utilized a well-known Bengali dataset of 11,807 comments to identify positive or negative sentiments. They used five cutting-edge transformer-based pre-trained models, including XLM-RoBERTa, BanglaBERT, Bangla-Bert-Base, DistilmBERT, and mBERT, with hyperparameter adjustment. Then, compared to the most current approaches used in the Bengali sentiment analysis problem, they suggest a combined model called Transformer-ensemble that exhibits exceptional detection performance with an accuracy of 95.97\% and an F1-score of 95.96\%. \textit{M.S. Islam et al.} \cite{islam2024sentiment} employed a new, extensive BangDSA dataset to examine several hybrid feature extraction methods and learning algorithms for Bangla sentiment analysis. They have used 21 distinct hybrid feature extraction techniques, such as GloVe, Word2Vec, N-gram, TF-IDF, FastText, and Bangla-BERT etc. In machine learning (ML), ensemble learning (EL), and deep learning (DL) methods, the suggested innovative method (Bangla-BERT+Skipgram), skipBangla-BERT, performs better than any other feature extraction tool. With an accuracy of 95.71\%, the CNN-BiLSTM hybrid approach outperforms the others. \textit{K. Ashraf et al.} \cite{ashraf} tried to clarify the elements that go into the prediction using Explainable AI and close the gap between the identification of abusive Bangla comments using NLP and five distinct ML classifiers, including DT, RF, Multinominal NB, SVM, and LR. They made use of n-gram TF-IDF representations. Using the Grid Search Cross Validation approach, the best set of hyperparameters for the machine learning models was discovered. With an accuracy of 85.57\%, Logistic Regression outperformed all other models. \textit{S.N. Nobel et al.} \cite{nobel} proposed a model that combines a hybrid CNN and LSTM for sentiment analysis For languages with limited resources, like Bengali. This study employed a dataset of 50,000 Bengali comments to train and test their system; they meticulously created the handcrafted dataset and made sure it was relevant and of high quality, they  concentrated on classifying comments into five categories based on their emotional content: threat, sexual, religious, troll, and not bully. The Hybrid CNN LSTM model obtains an exceptional accuracy rate of 92.09\%, outperforming traditional methods. \textit{T. Samin et al.} \cite{samin2024sentiment} proposed a model to do sentiment analysis on news stories that appear on the front pages of well-known Bangla newspapers. The polarity of the entire dataset as well as the specific polarity of news articles from five different newspapers were investigated. The dataset contains three types of sentiment: positive, negative, and neutral. Six classification models, LR, Multinomial NB, SVM, SVC, RF, and KNN, were used for the classification and analysis. Achieving an accuracy of 85.19\% other models were outperformed by logistic regression. \textit{S. Sunny et al.} \cite{sunny2024bangla} investigate the opinions expressed in Bangla customer reviews and offer insightful information to prospective customers. The suggested methodology entails a thorough approach to sentiment identification and data collection from online purchasing platforms. For efficient feature selection, TF-IDF was combined with trigram features. NB, RF, Extra Tree, KNN, LR, SVM, Artificial Neural Network, and Multilayer Perceptron were among the classifiers used to classify attitudes. Using the suggested methods, the Multilayer Perceptron model demonstrated exceptional performance in evaluating sentiments about Bangla products, achieving an accuracy rate of 96.42\%. \textit{F. Khanam et al.} \cite{khanam2024bangla} proposed a model to identify sentiments on Bangla news items. They tested a variety of models, including CNN, LSTM, the transformer model like Bangla-BERT-base, and deep learning models, in addition to conventional machine learning models. The Bangla-BERT base, using 10-fold cross-validation, attained a noteworthy 96\% accuracy outperform other models. \vspace{-2mm}

\section{MATERIALS \& METHODS} 
\begin{figure}[htbp]
	\centering 
	\fbox{\includegraphics[height=10cm, width=8.5cm]{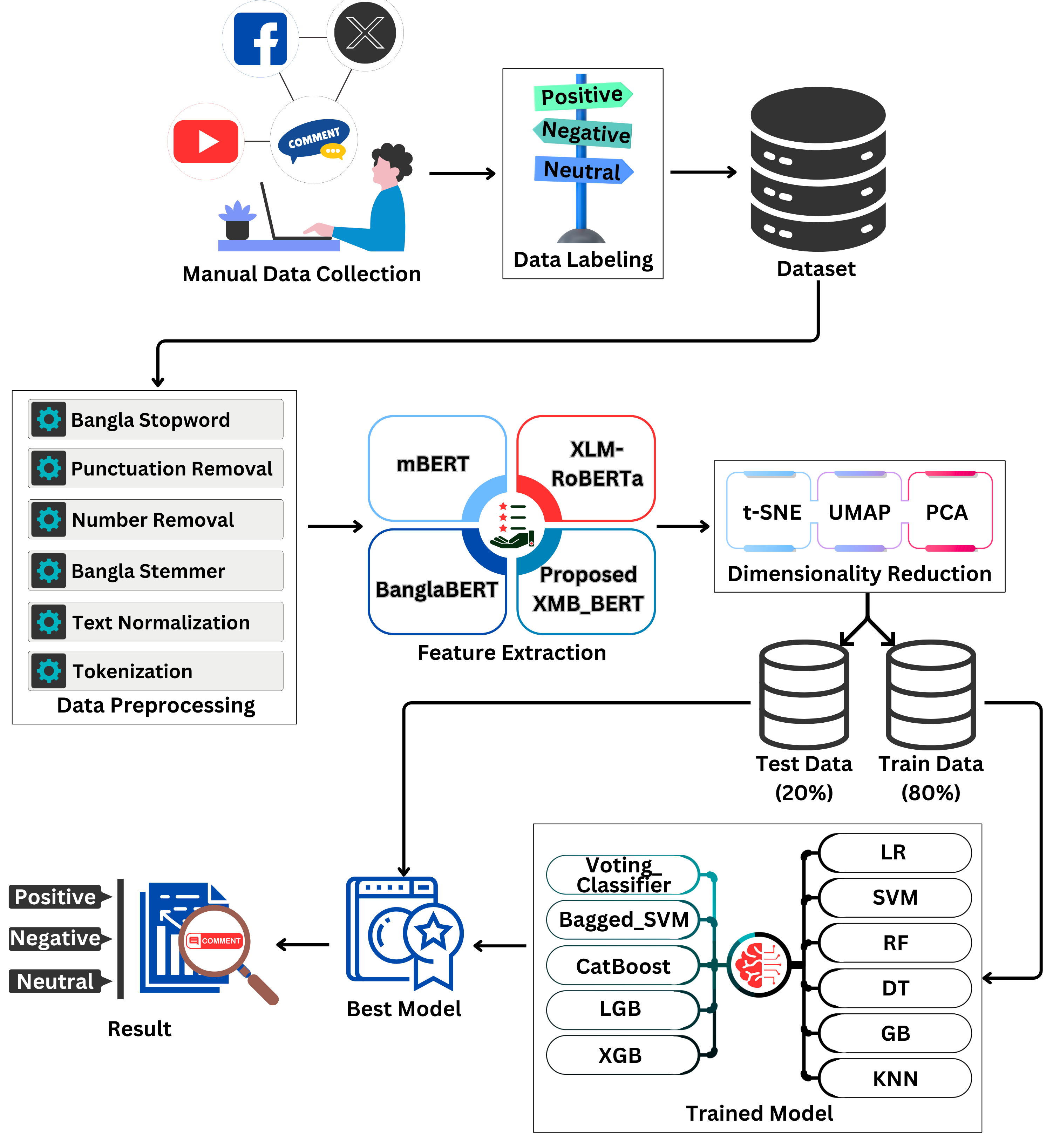}}
		\caption{Proposed Workflow Diagram for Sentiment Analysis} 
	\label{fig:pmd}
\end{figure}

\vspace{-2mm} This section will go over the approach taken for sentiment analysis of the public comments in social media during and after the student protest in Bangladesh. It includes a number of procedures, including gathering data, preprocessing, feature extraction, dimensionality reduction, and machine learning model selection.  Figure \ref{fig:pmd} displays the proposed workflow diagram.
 \vspace{-2mm}

\subsection{Data Collection}
\begin{figure}[]
	\centering 
	\includegraphics[height=5.1cm, width=8.5cm]{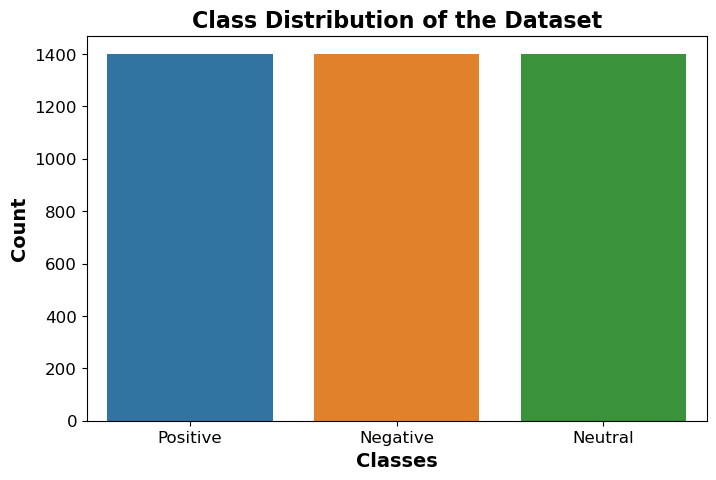}
		\caption{Class Distribution of the Dataset across three sentiment classes: Positive, Negative, and Neutral. The counts for each class are balanced. \vspace{-2mm}} 
	\label{fig:cdlb}
\end{figure}

\begin{figure}[htbp]
	\centering 
	\includegraphics[height=10.75cm, width=8.75cm]{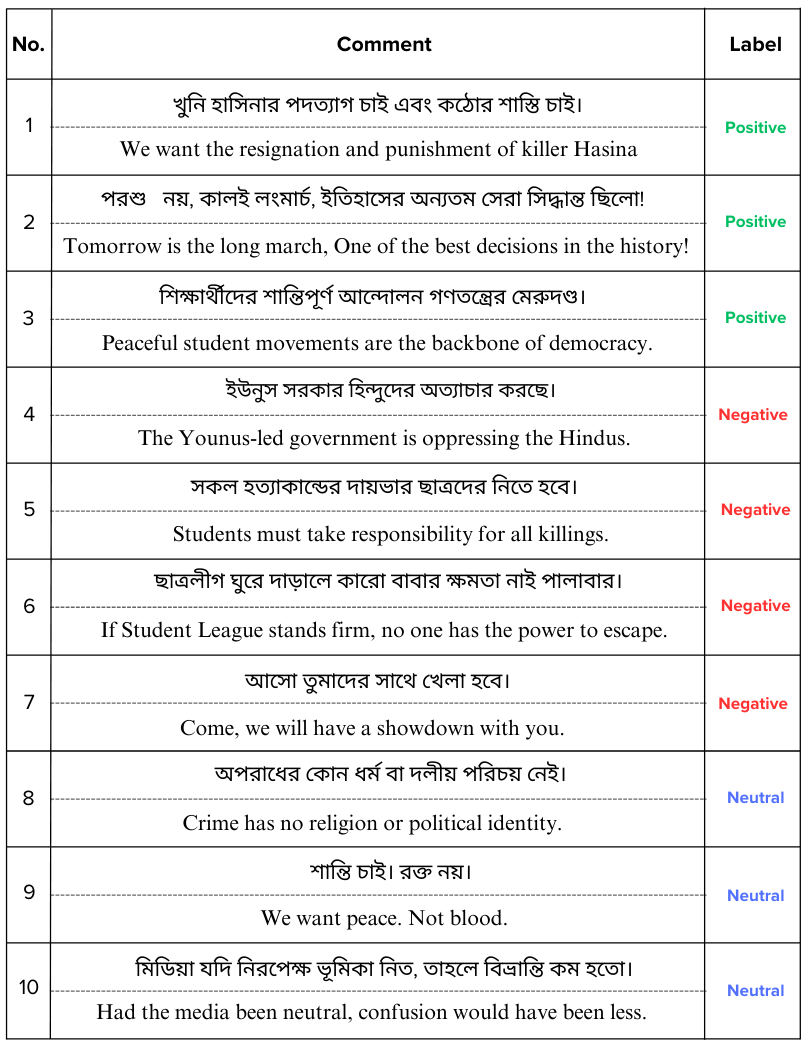}
		\caption{Sample comments from the collected dataset, along with their English translations and labels, indicating positive, negative or neutral comments.} 
	\label{fig:tables}
\end{figure}

The data for this sentiment analysis study was meticulously curated from comments collected during and after the July Revolution from three popular social media platforms: Facebook, YouTube, and Twitter (X). A total of 4,200 Bangla-language comments were gathered manually, encompassing sentiments classified as positive, negative, and neutral, with each sentiment class comprising an equal distribution of 1,400 comments to ensure a balanced dataset. Fig. \ref{fig:cdlb} shows the bar chart of the class distribution of the dataset. The dataset was collected between September and October by searching posts and news related to the July Revolution. The collected data was labeled based on specific criteria: comments that expressed support for the protest or included positive remarks were marked as positive, comments containing slang words or threats directed at the protesting students were categorized as negative, and comments with a neutral tone that neither supported the students nor used offensive language were labeled as neutral. The finalized dataset was structured into a CSV file to train the proposed model. We named the dataset "Student-People Mass Uprising Public Sentiments Dataset." Fig. \ref{fig:tables} presents a selection of sample comments from the collected dataset, along with their English translations.

\begin{figure}[]
	\centering 
	\includegraphics[height=5.1cm, width=8.5cm]{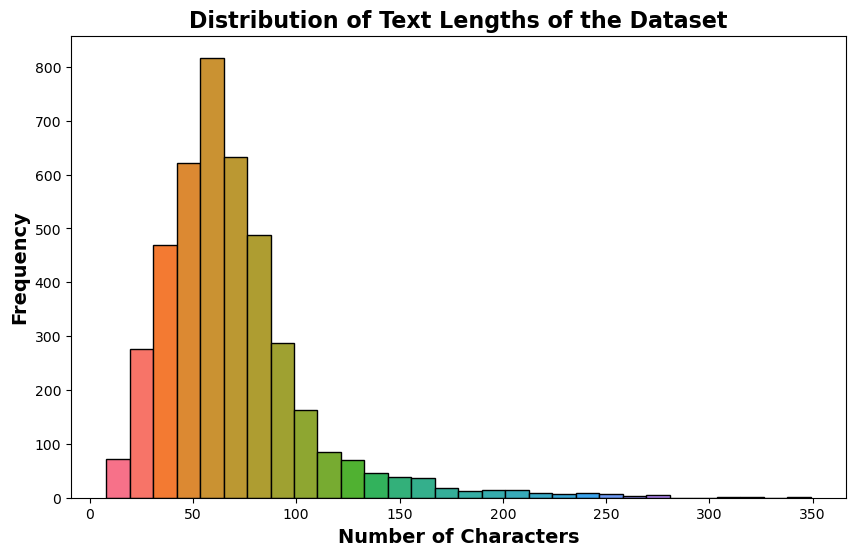}
		\caption{Distribution of text lengths in the dataset, indicating a right-skewed pattern with most samples concentrated around 50-60 characters. \vspace{-2mm}} 
	\label{fig:cdl}
\end{figure}

\noindent Fig. \ref{fig:cdl} presents the histogram illustrating the distribution of text lengths in our dataset, measured in the number of characters. The x-axis represents the number of characters in the text samples, while the y-axis denotes their frequency. The distribution appears right-skewed, with most text samples having a length between 40 and 100 characters, peaking around 50-60 characters. As the character length increases, the frequency of occurrences declines significantly, with very few instances exceeding 200 characters. The histogram is color-coded in a gradient, providing a visual emphasis on the density of occurrences across different text lengths. \vspace{2mm}

\subsection{Data Preprocessing}
In the data preprocessing phase, we applied a series of essential techniques to clean and prepare the dataset for sentiment analysis. The dataset does not contain any null value. First, tokenization was performed to split the text into individual words or tokens. Next, Bangla stopwords—common words that do not carry significant meaning were removed to focus on meaningful content. Punctuation marks and numbers were also eliminated to reduce noise in the data. Additionally, a Bangla stemmer was employed to reduce words to their root forms, ensuring that different inflections of the same word were treated uniformly. Finally, text normalization was carried out to standardize variations in spelling and formatting, creating a cleaner and more consistent dataset for model training and evaluation. This comprehensive preprocessing pipeline was crucial for optimizing the performance of our sentiment analysis model. \vspace{2mm}

\begin{figure*}[htbp]
    \centering
    \subfigure[]{\includegraphics[height=3cm, width=4.375cm]{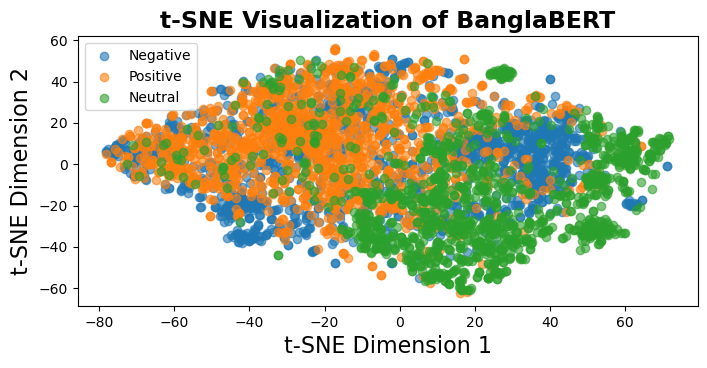}}
    \subfigure[]{\includegraphics[height=3cm, width=4.375cm]{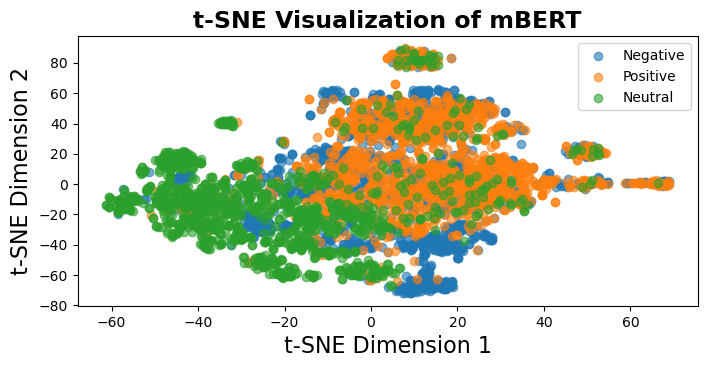}}
    \subfigure[]{\includegraphics[height=3cm, width=4.375cm]{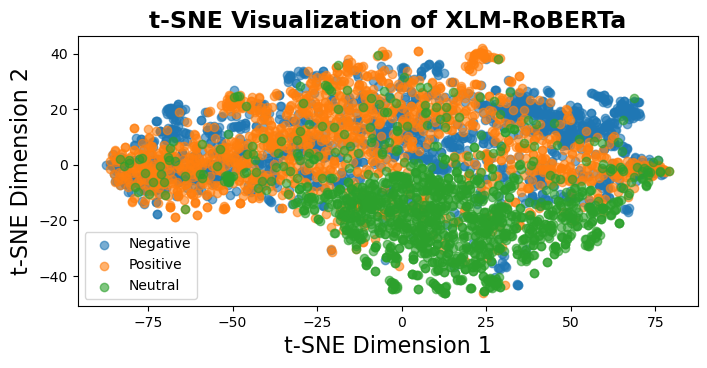}}
    \subfigure[]{\includegraphics[height=3cm, width=4.375cm]{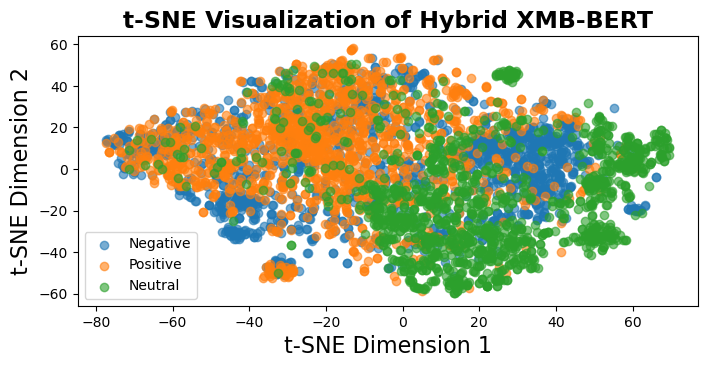}}
    \vspace{-0.5cm}
    \caption{t-distributed Stochastic Neighbor Embedding (t-SNE) visualize the data before dimensionality reduction. (a) t-SNE visualization of BanglaBERT, (b) t-SNE visualization of mBERT, (c) t-SNE visualization of XLM-RoBERTa, and (d) t-SNE visualization of proposed hybrid XMB-BERT.} \vspace{-2mm}
    \label{fig:tsne}
\end{figure*}
\begin{figure*}[htbp]
    \centering
    \subfigure[]{\includegraphics[height=3cm, width=4.375cm]{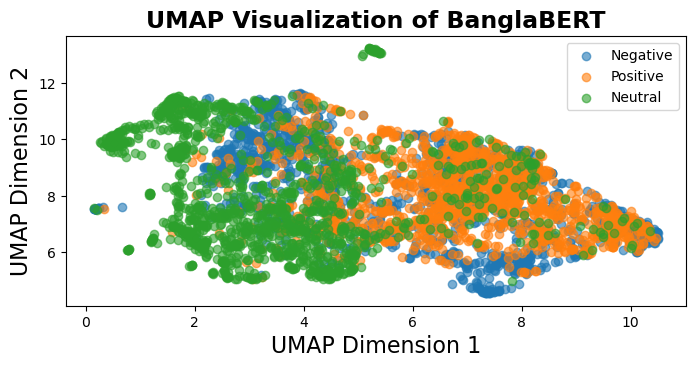}}
    \subfigure[]{\includegraphics[height=3cm, width=4.375cm]{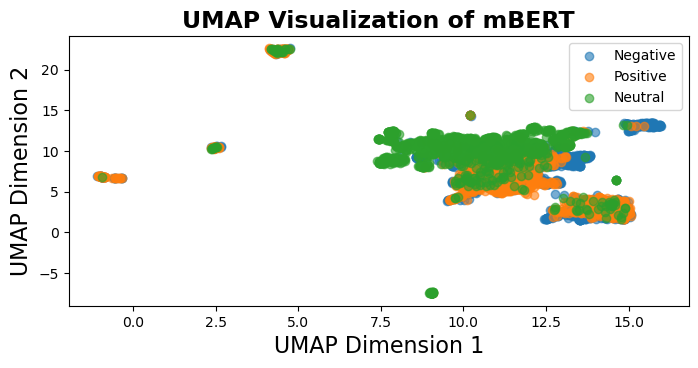}}
    \subfigure[]{\includegraphics[height=3cm, width=4.375cm]{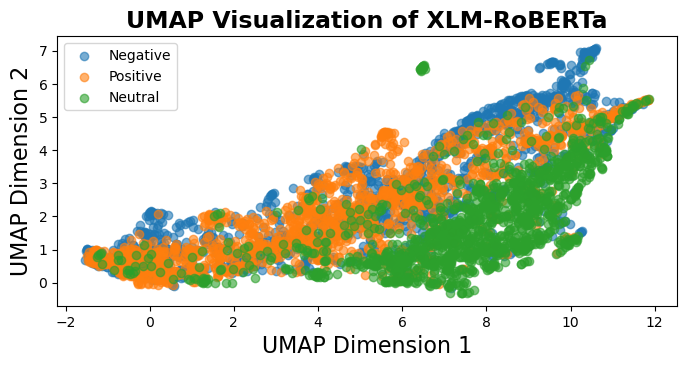}}
    \subfigure[]{\includegraphics[height=3cm, width=4.375cm]{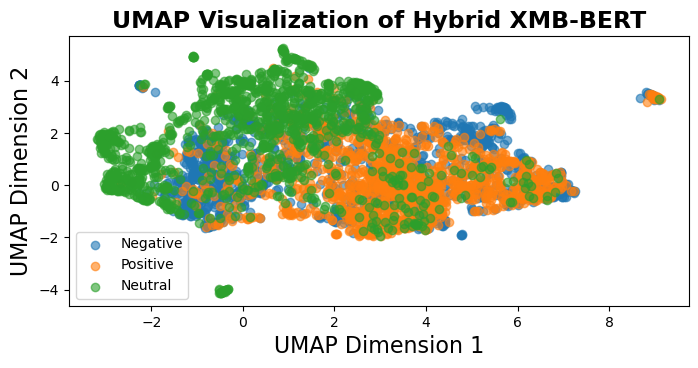}}
    \vspace{-0.5cm}
    \caption{Uniform Manifold Approximation and Projection (UMAP) visualize data before dimensionality reduction. (a) UMAP visualization of BanglaBERT, (b) UMAP visualization of mBERT, (c) UMAP visualization of XLM-RoBERTa, and (d) UMAP visualization of proposed XMB-BERT.} \vspace{-2mm}
    \label{fig:umap}
\end{figure*}
\begin{figure*}[htbp]
    \centering
    \subfigure[]{\includegraphics[height=3cm, width=4.375cm]{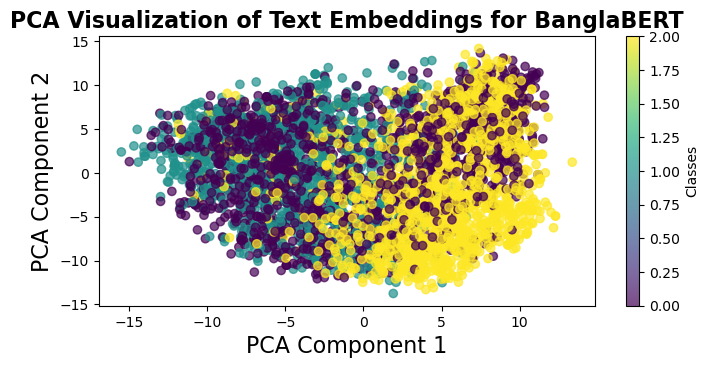}}
    \subfigure[]{\includegraphics[height=3cm, width=4.375cm]{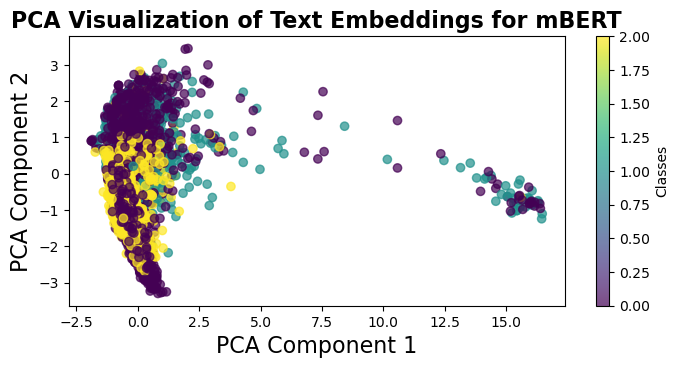}}
    \subfigure[]{\includegraphics[height=3cm, width=4.375cm]{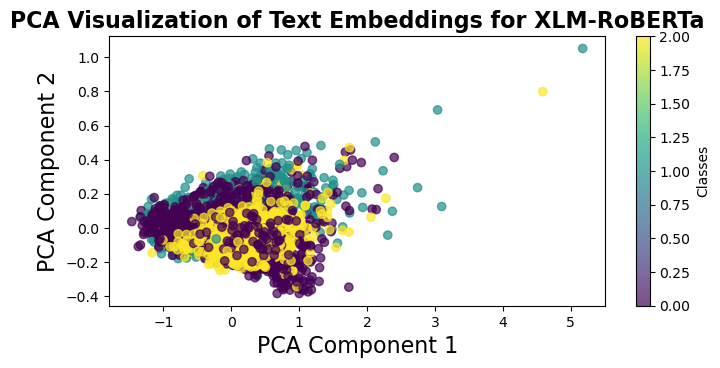}}
    \subfigure[]{\includegraphics[height=3cm, width=4.375cm]{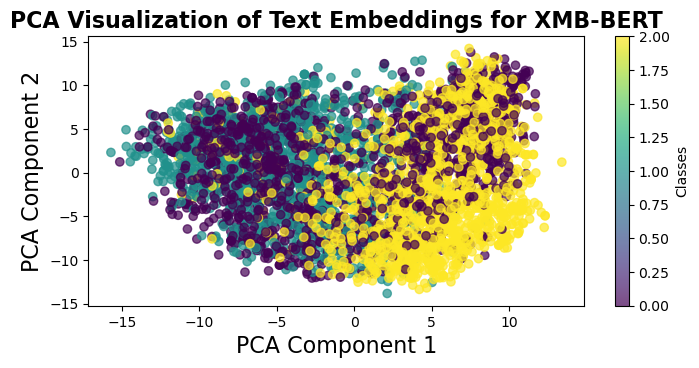}}
    \vspace{-0.5cm}
    \caption{Principle Component Analysis (PCA) visualizes the high dimensional data after dimensionality reduction. (a) PCA visualization of BanglaBERT, (b) PCA visualization of mBERT, (c) PCA visualization of XLM-RoBERTa, and (d) PCA visualization of proposed hybrid XMB-BERT.} \vspace{-0.4cm}
    \label{fig:pca}
\end{figure*}

\subsection{Data Partitioning}
To effectively train and evaluate our proposed sentiment classification models, we partitioned the Student-People Mass Uprising Public Sentiments Dataset into training and testing subsets. As shown in Table \ref{tab:data},  
\begin{table}[h]
\caption{Student-People Mass Uprising Public Sentiments Dataset}
\centering
\setlength{\tabcolsep}{16pt}
\renewcommand{\arraystretch}{1}
\begin{tabular}{@{}cccc@{}}
\toprule
\textbf{Sentiments} & \textbf{Training} & \textbf{Testing} & \textbf{Total Samples} \\ \hline
Negative & 1120               & 280               & 1400                   \\
Positive  & 1120               & 280               & 1400                   \\ 
Neutral  & 1120              & 280               & 1400   \\ 

\textbf{Total} & \textbf{3360} & \textbf{840} & \textbf{4200} \\
\bottomrule
\end{tabular}
\label{tab:data}
\end{table}
the dataset comprises three sentiment categories: Negative, Positive, and Neutral, each containing a total of 1400 samples. These samples were evenly divided into an 80:20 train and test ratio, 1120 instances for training and 280 instances for testing across each sentiment class. This uniform distribution ensures a balanced dataset, which is essential for reducing class bias during model training and evaluation. By maintaining equal class representation, the models are encouraged to learn features from all sentiment categories with equal emphasis, thereby improving classification fairness and performance across classes.

\subsection{Feature Extraction \& Proposed Hybrid Model}
\begin{figure*}[htbp]
	\centering 
	\includegraphics[height=4.6cm, width=17.5cm]{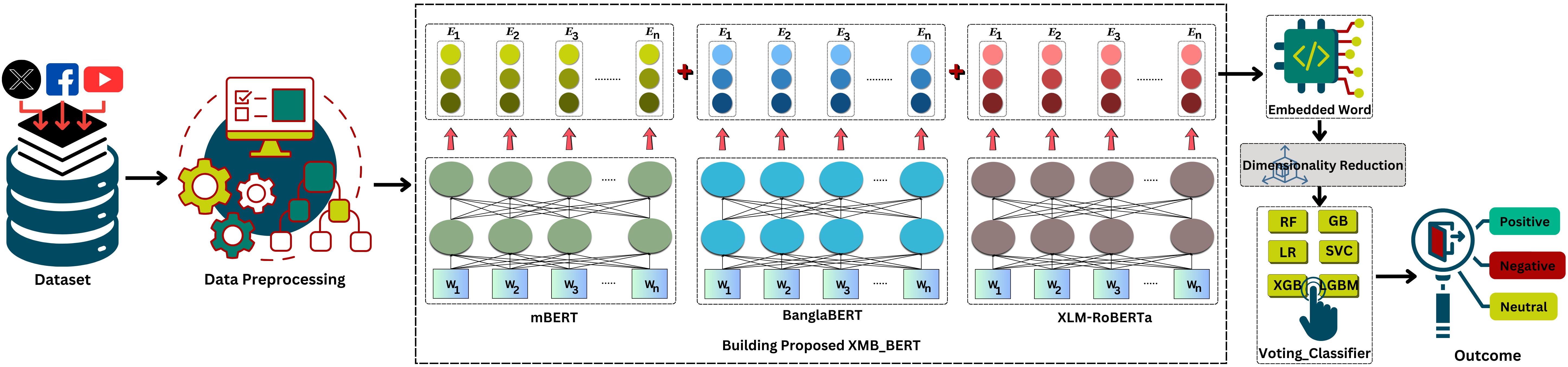}
		\caption{The Architecture of the proposed Hybrid XMB-BERT. It shows that preprocessed data is fed into three transformer models: mBERT, BanglaBERT, and XLM-RoBERTa, to generate contextual embeddings. These embeddings are then combined to construct the hybrid XMB-BERT architecture, which effectively captures diverse linguistic features. Dimensionality reduction is applied to the combined feature representations to simplify and streamline the data. Finally, the reduced features are classified using an ensemble voting classifier to perform sentiment analysis.} \vspace{-3mm}
	\label{fig:xmb}
\end{figure*}

For feature extraction, we employed three state-of-the-art transformer-based language models: mBERT \cite{mbert}, XLM-Roberta \cite{XLM}, and Bangla-BERT \cite{banglabert}. Each model was utilized separately to generate contextualized embeddings for the textual data, capturing semantic and syntactic relationships specific to their respective training paradigms. Specifically, we extracted the embeddings corresponding to the [CLS] token, which represents a sentence-level summary of the input text. To enhance the representation and leverage the complementary strengths of these models, we developed a customized feature extraction approach by combining the embeddings from all three models, and we named this model as Hybrid XMB-BERT; for this, we took the first letter of each model, X from XLM-RoBERTa, m from mBERT and B from Bangla BERT, and finally, we named it XMB-BERT. This was achieved through concatenation, resulting in a unified feature vector that integrates the multilingual capabilities of mBERT, the Bangla-specific contextual understanding of Bangla-BERT, and the cross-lingual proficiency of XLM-Roberta. This comprehensive feature extraction process allowed us to create a robust representation of the text, optimized for downstream analysis and classification. Figure \ref{fig:xmb} represents the architecture of the proposed XMB-BERT.

\subsection{Dimensionality Reduction}
We applied t-distributed Stochastic Neighbor Embedding (t-SNE) \cite{tsne} and Uniform Manifold Approximation and Projection (UMAP) \cite{umap} for visualization, and used Principal Component Analysis (PCA) \cite{pca} for dimensionality reduction. t-SNE and UMAP helped in visualizing the data distribution and uncovering patterns in the feature space, revealing the separability and clustering of different sentiment classes. PCA was then employed to reduce the feature dimensionality while preserving important variance, thus enhancing the feature representation for downstream machine learning tasks. The comparative results from these methods are shown in Figures \ref{fig:tsne}, \ref{fig:umap}, and \ref{fig:pca}, demonstrating how this multi-step approach refines the feature space. These visualizations not only provided a 
clearer understanding of the data’s inherent structure but also contributed to improved model performance by aligning the features with sentiment categories. \vspace{3mm}

\subsection{Model Selection}
We fed the extracted features after reducing dimensionality to 11 machine learning algorithms: Logistic Regression (LR) \cite{lr}, Support Vector Machines (SVM) \cite{svm_mostcited} \cite{strawberry}, Decision Trees (DT) \cite{dt}, Random Forests (RF) \cite{rf1}, k-nearest Neighbors (KNN) \cite{knn} \cite{sourov}, Extreme Gradient Boosting (XGBoost) \cite{xgb} \cite{maternal}, Light Gradient Boosting Machine (LGBM) \cite{lgbm}, CatBoost \cite{catboost}, Bagged SVM \cite{baggingsvm}, Voting Classifier  \cite{hossen2025efficientdeeplearningframework} \cite{voting}, and Gradient Boosting \cite{gb} \cite{jelly} for sentiment analysis on social media comments on the July Revolution. Each algorithm was trained and tested on the extracted feature representations to classify the sentiments effectively. This comprehensive evaluation allowed us to identify the most effective model for sentiment analysis, leveraging both individual classifier performance and ensemble learning techniques to achieve robust and accurate results.

\section{EXPERIMENTAL SETUP \& IMPLEMENTATION}
\subsection{Experimental Setup}
The experiment was conducted using a Windows 11 operating system on a ×64 processor, intel(R) core i5 8th Gen CPU running at 1.60GHz to 3.90GHz with 12 GB of DDR4 RAM and NVIDIA MX250 (2 GB V-Ram) GPU. The Jupyter Notebook was used in conjunction with Anaconda Navigator for applications involving natural language processing. It takes more than three hours to train and evaluate the models.

\subsection{Implementation Details}

The implementation procedure for the sentiment analysis follows a systematic and modular approach as outlined in the algorithm \ref{alg:sentiment}. First, the dataset is initialized, and preprocessing techniques were applied to clean the data. Feature extraction is then performed using various methods like BanglaBERT, mBERT, XLM-RoBERTa, and proposed Hybrid XMB-BERT to generate meaningful vector representations. Dimensionality reduction and visualization techniques, including t-SNE, UMAP, and PCA were employed to visualize and reduce feature complexity. The data is divided into 80:20 training and testing ratios for each feature extraction and dimensionality reduction technique combination. A range of advanced and traditional classifiers are trained on these transformed features. Performance metrics like accuracy, precision, recall, and F1-score are recorded after predictions. This iterative process systematically evaluates the effectiveness of various combinations, allowing for the selection of the most suitable model for sentiment classification in three classes.

\begin{algorithm}[]
\scriptsize
\caption{Algorithm for Sentiment Analysis}
\label{alg:sentiment}

\textbf{Initialize:} $DS$, $PP$, $FE$, $DR$, $NC$ \tcp*[r]{Initialize components}

$DS \gets \text{Data Set}, \; PP \gets \text{Pre-Processing}$ \;
$FE \gets \text{Feature Extraction}, \; DR \gets \text{Dimensionality Reduction}$ \;
$CLF \gets \text{Classifiers}$ \;
$PP = [\text{Stop Words, Punctuation Removal, Tokenization, Stemming}]$ \;
$FE = [\text{mBERT, BanglaBERT, XLM-RoBERTa, Hybrid XMB-BERT}]$ \; 
$DR = [\text{t-SNE, PCA, UMAP}]$ \; 
$NC = [\text{LR, SVM, RF, DT, KNN, XGB, LGBM, CatBoost, BSVM, VC}]$ \;

$X \gets DS[\text{Content}], \; Y \gets DS[\text{Label]}$ \;

\For{$i \gets 1$ \KwTo $|FE|$}{
    $XV \gets FE[i](X)$ \tcp*[r]{Generate Word Embeddings}
    \For{$j \gets 1$ \KwTo $|DR|$}{
        $X_{DR} \gets DR[j](XV)$ \tcp*[r]{Apply Dimensionality Reduction}
        \For{$k \gets 1$ \KwTo $|NC|$}{
            $X_T, Y_T, x_t, y_t \gets \text{Train-Test Split (80:20)}(X_{DR}, Y)$ \;
            $NC[k].\text{fit}(X_T, Y_T)$ \tcp*[r]{Train Classifier}
            $y_{\text{pred}} \gets NC[k].\text{predict}(x_t)$ \tcp*[r]{Test Classifier}
            Display $\to [\text{Accuracy, Precision, Recall, F1-Score}]$ \;
        }
    }
}
\textbf{Deinitialize:} $DS$, $PP$, $FE$, $DR$, $NC$ \tcp*[r]{Shutdown components}

\end{algorithm}

\section{RESULT ANALYSIS \& DISCUSSION}
This section presents a comprehensive analysis of the performance of various sentiment analysis models applied to the collected dataset. We evaluate the models using multiple performance metrics such as accuracy, precision, recall, and F1-score, offering insights into their efficacy in classifying sentiments. The findings highlight the effectiveness of different feature extraction techniques, classifiers, and hybrid approaches, providing a comparative perspective to determine the optimal configuration for sentiment analysis in Bangla-language comments. The results are visualized through confusion matrices, ROC curves, and detailed tabular comparisons, illustrating the strengths and limitations of each method.

\begin{table*}[]
\centering
\scriptsize
\setlength{\tabcolsep}{5.3pt}
\renewcommand{\arraystretch}{1.2895}
\caption{Comparison of Accuracy, Precision, Recall, and F1-Score of Different Feature Extraction Techniques and Classifiers}
\begin{tabular}{|c|cccccccccccccccc|}
\hline
\multirow{3}{*}{\textbf{Classifier}} & \multicolumn{16}{c|}{\textbf{Feature Extraction Techniques}} \\ \cline{2-17} 
 & \multicolumn{4}{c|}{\textbf{BanglaBERT}} & \multicolumn{4}{c|}{\textbf{mBERT}} & \multicolumn{4}{c|}{\textbf{XLM-RoBERTa}} & \multicolumn{4}{c|}{\textbf{XMB\_BERT}} \\ \cline{2-17} 
 & \multicolumn{1}{c|}{\textbf{A(\%)}} & \multicolumn{1}{c|}{\textbf{P(\%)}} & \multicolumn{1}{c|}{\textbf{R(\%)}} & \multicolumn{1}{c|}{\textbf{F1(\%)}} & \multicolumn{1}{c|}{\textbf{A(\%)}} & \multicolumn{1}{c|}{\textbf{P(\%)}} & \multicolumn{1}{c|}{\textbf{R(\%)}} & \multicolumn{1}{c|}{\textbf{F1(\%)}} & \multicolumn{1}{c|}{\textbf{A(\%)}} & \multicolumn{1}{c|}{\textbf{P(\%)}} & \multicolumn{1}{c|}{\textbf{R(\%)}} & \multicolumn{1}{c|}{\textbf{F1(\%)}} & \multicolumn{1}{c|}{\textbf{A(\%)}} & \multicolumn{1}{c|}{\textbf{P(\%)}} & \multicolumn{1}{c|}{\textbf{R(\%)}} & \textbf{F1(\%)} \\ \hline
LR & \multicolumn{1}{c|}{74.7} & \multicolumn{1}{c|}{74.7} & \multicolumn{1}{c|}{74.4} & \multicolumn{1}{c|}{74.5} & \multicolumn{1}{c|}{76.3} & \multicolumn{1}{c|}{77.4} & \multicolumn{1}{c|}{76.5} & \multicolumn{1}{c|}{77.3} & \multicolumn{1}{c|}{78.1} & \multicolumn{1}{c|}{77.3} & \multicolumn{1}{c|}{78.1} & \multicolumn{1}{c|}{77.2} & \multicolumn{1}{c|}{75.6} & \multicolumn{1}{c|}{75.6} & \multicolumn{1}{c|}{75.6} & 75.8 \\ \hline
SVM & \multicolumn{1}{c|}{73.0} & \multicolumn{1}{c|}{75.0} & \multicolumn{1}{c|}{73.5} & \multicolumn{1}{c|}{73.4} & \multicolumn{1}{c|}{75.4} & \multicolumn{1}{c|}{76.4} & \multicolumn{1}{c|}{76.3} & \multicolumn{1}{c|}{76.2} & \multicolumn{1}{c|}{78.2} & \multicolumn{1}{c|}{79.3} & \multicolumn{1}{c|}{78.4} & \multicolumn{1}{c|}{78.2} & \multicolumn{1}{c|}{74.1} & \multicolumn{1}{c|}{75.2} & \multicolumn{1}{c|}{75.0} & 75.0 \\ \hline
RF & \multicolumn{1}{c|}{77.4} & \multicolumn{1}{c|}{78.4} & \multicolumn{1}{c|}{77.3} & \multicolumn{1}{c|}{77.3} & \multicolumn{1}{c|}{75.6} & \multicolumn{1}{c|}{76.0} & \multicolumn{1}{c|}{75.5} & \multicolumn{1}{c|}{75.5} & \multicolumn{1}{c|}{78.1} & \multicolumn{1}{c|}{79.1} & \multicolumn{1}{c|}{78.1} & \multicolumn{1}{c|}{78.1} & \multicolumn{1}{c|}{79.4} & \multicolumn{1}{c|}{80.1} & \multicolumn{1}{c|}{79.4} & 79.4 \\ \hline
DT & \multicolumn{1}{c|}{60.9} & \multicolumn{1}{c|}{60.9} & \multicolumn{1}{c|}{60.9} & \multicolumn{1}{c|}{60.9} & \multicolumn{1}{c|}{61.1} & \multicolumn{1}{c|}{62.0} & \multicolumn{1}{c|}{61.1} & \multicolumn{1}{c|}{61.1} & \multicolumn{1}{c|}{65.4} & \multicolumn{1}{c|}{66.0} & \multicolumn{1}{c|}{65.4} & \multicolumn{1}{c|}{65.4} & \multicolumn{1}{c|}{63.2} & \multicolumn{1}{c|}{64.2} & \multicolumn{1}{c|}{64.2} & 64.2 \\ \hline
GB & \multicolumn{1}{c|}{75.1} & \multicolumn{1}{c|}{75.1} & \multicolumn{1}{c|}{75.1} & \multicolumn{1}{c|}{75.1} & \multicolumn{1}{c|}{76.3} & \multicolumn{1}{c|}{76.3} & \multicolumn{1}{c|}{76.3} & \multicolumn{1}{c|}{76.3} & \multicolumn{1}{c|}{80.5} & \multicolumn{1}{c|}{80.4} & \multicolumn{1}{c|}{80.5} & \multicolumn{1}{c|}{80.4} & \multicolumn{1}{c|}{82.3} & \multicolumn{1}{c|}{82.1} & \multicolumn{1}{c|}{82.2} & 82.3 \\ \hline
KNN & \multicolumn{1}{c|}{75.4} & \multicolumn{1}{c|}{75.3} & \multicolumn{1}{c|}{75.3} & \multicolumn{1}{c|}{75.4} & \multicolumn{1}{c|}{73.3} & \multicolumn{1}{c|}{74.2} & \multicolumn{1}{c|}{74.2} & \multicolumn{1}{c|}{74.2} & \multicolumn{1}{c|}{74.5} & \multicolumn{1}{c|}{74.1} & \multicolumn{1}{c|}{74.3} & \multicolumn{1}{c|}{74.4} & \multicolumn{1}{c|}{73.4} & \multicolumn{1}{c|}{73.5} & \multicolumn{1}{c|}{74.2} & 73.7 \\ \hline
XGB & \multicolumn{1}{c|}{78.4} & \multicolumn{1}{c|}{78.4} & \multicolumn{1}{c|}{78.4} & \multicolumn{1}{c|}{78.4} & \multicolumn{1}{c|}{77.8} & \multicolumn{1}{c|}{78.5} & \multicolumn{1}{c|}{78.0} & \multicolumn{1}{c|}{77.8} & \multicolumn{1}{c|}{81.6} & \multicolumn{1}{c|}{82.5} & \multicolumn{1}{c|}{81.6} & \multicolumn{1}{c|}{82.0} & \multicolumn{1}{c|}{82.6} & \multicolumn{1}{c|}{83.4} & \multicolumn{1}{c|}{82.8} & 82.8 \\ \hline
LGBM & \multicolumn{1}{c|}{77.5} & \multicolumn{1}{c|}{78.2} & \multicolumn{1}{c|}{77.5} & \multicolumn{1}{c|}{77.5} & \multicolumn{1}{c|}{\textbf{78.4}} & \multicolumn{1}{c|}{\textbf{79.5}} & \multicolumn{1}{c|}{\textbf{78.5}} & \multicolumn{1}{c|}{\textbf{78.5}} & \multicolumn{1}{c|}{\textbf{82.2}} & \multicolumn{1}{c|}{\textbf{83.1}} & \multicolumn{1}{c|}{\textbf{82.2}} & \multicolumn{1}{c|}{\textbf{82.2}} & \multicolumn{1}{c|}{82.5} & \multicolumn{1}{c|}{83.1} & \multicolumn{1}{c|}{82.5} & 82.5 \\ \hline
CatBoost & \multicolumn{1}{c|}{75.3} & \multicolumn{1}{c|}{75.3} & \multicolumn{1}{c|}{75.3} & \multicolumn{1}{c|}{74.2} & \multicolumn{1}{c|}{75.3} & \multicolumn{1}{c|}{76.1} & \multicolumn{1}{c|}{75.3} & \multicolumn{1}{c|}{75.1} & \multicolumn{1}{c|}{78.5} & \multicolumn{1}{c|}{79.2} & \multicolumn{1}{c|}{78.5} & \multicolumn{1}{c|}{78.6} & \multicolumn{1}{c|}{80.2} & \multicolumn{1}{c|}{80.5} & \multicolumn{1}{c|}{79.9} & 80.1 \\ \hline
Bagged\_SVM & \multicolumn{1}{c|}{\textbf{81.4}} & \multicolumn{1}{c|}{\textbf{82.0}} & \multicolumn{1}{c|}{\textbf{81.9}} & \multicolumn{1}{c|}{\textbf{81.8}} & \multicolumn{1}{c|}{76.9} & \multicolumn{1}{c|}{77.1} & \multicolumn{1}{c|}{76.9} & \multicolumn{1}{c|}{76.9} & \multicolumn{1}{c|}{49.9} & \multicolumn{1}{c|}{47.9} & \multicolumn{1}{c|}{47.9} & \multicolumn{1}{c|}{49.9} & \multicolumn{1}{c|}{81.6} & \multicolumn{1}{c|}{82.1} & \multicolumn{1}{c|}{81.1} & 81.3 \\ \hline
Voting\_Classifier & \multicolumn{1}{c|}{80.5} & \multicolumn{1}{c|}{80.4} & \multicolumn{1}{c|}{80.5} & \multicolumn{1}{c|}{80.3} & \multicolumn{1}{c|}{77.3} & \multicolumn{1}{c|}{78.2} & \multicolumn{1}{c|}{77.5} & \multicolumn{1}{c|}{77.5} & \multicolumn{1}{c|}{81.9} & \multicolumn{1}{c|}{82.5} & \multicolumn{1}{c|}{81.5} & \multicolumn{1}{c|}{81.5} & \multicolumn{1}{c|}{\textbf{83.7}} & \multicolumn{1}{c|}{\textbf{84.1}} & \multicolumn{1}{c|}{\textbf{83.7}} & \textbf{83.7} \\ \hline
\end{tabular}
\label{tab:compare}
\end{table*}

\begin{figure*}[htbp]
    \centering
    \subfigure[]{\label{fig:a}\includegraphics[height=3.9cm, width=4.375cm]{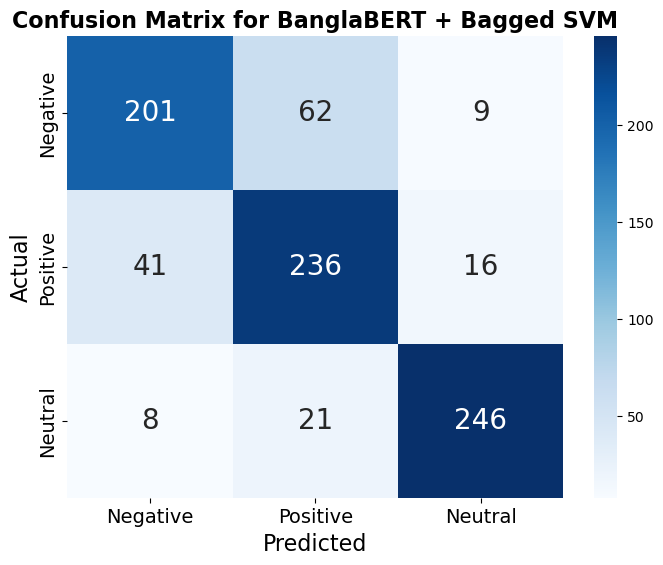}}
    \subfigure[]{\label{fig:b}\includegraphics[height=3.9cm, width=4.375cm]{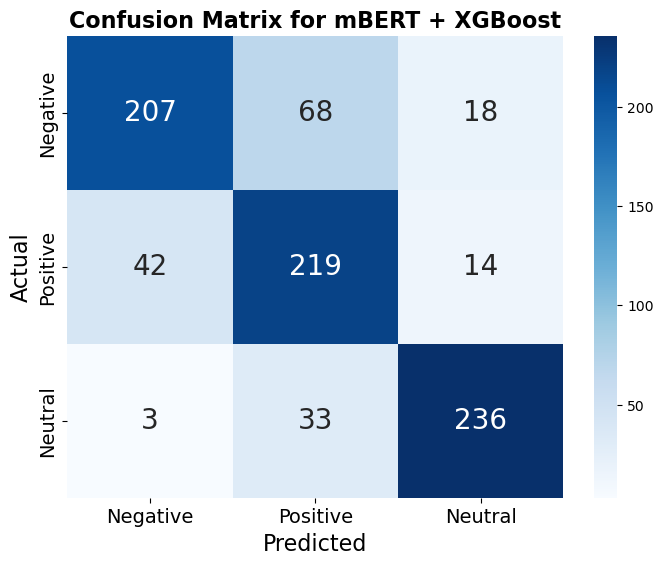}}
    \subfigure[]{\label{fig:c}\includegraphics[height=3.9cm, width=4.375cm]{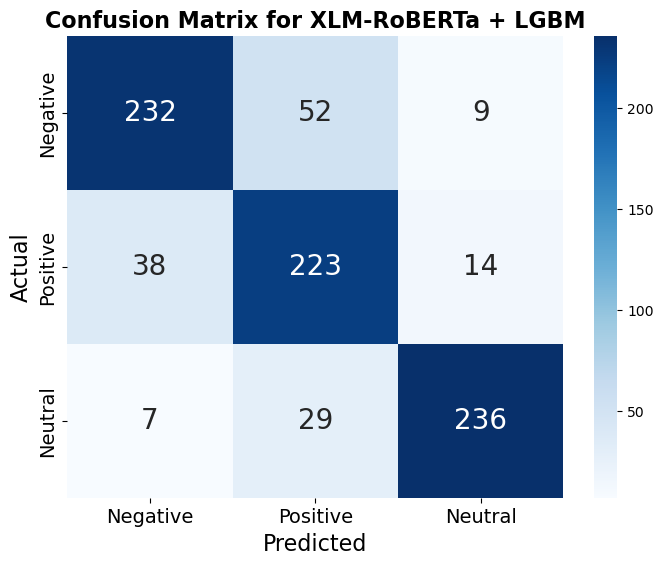}}
    \subfigure[]{\label{fig:d}\includegraphics[height=3.9cm, width=4.375cm]{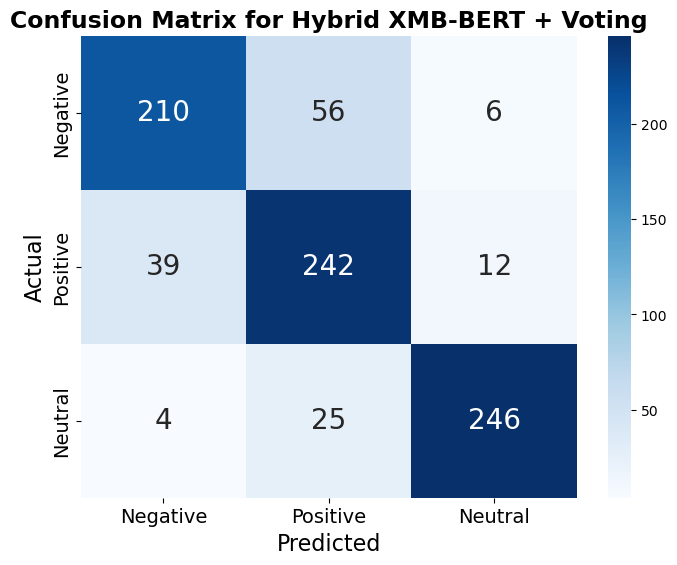}}
    \vspace{-0.5cm}
    \caption{Confusion matrices compare the performance of three different feature extraction techniques with Proposed Hybrid XMB-BERT: (a) BanglaBERT with Bagged SVM, (b) Multilingual BERT (mBERT) with Extreme Gradient Boosting (XGBoost, (c) XLM-RoBERTa with Light Gradient Boosting Machine (LGBM), and (d) Proposed Hybrid XMB-BERT with Voting Classifier,  shows the classification performance. }

    \label{fig:cm}
\end{figure*}

\begin{figure*}[htbp]
    \centering
    \subfigure[]{\label{fig:e}\includegraphics[height=3cm, width=4.375cm]{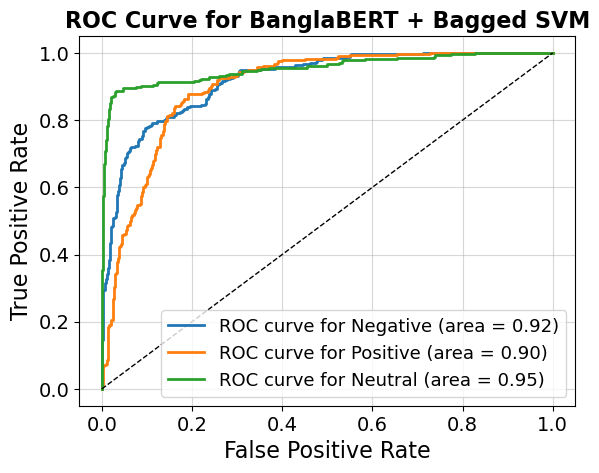}}
    \subfigure[]{\label{fig:f}\includegraphics[height=3cm, width=4.375cm]{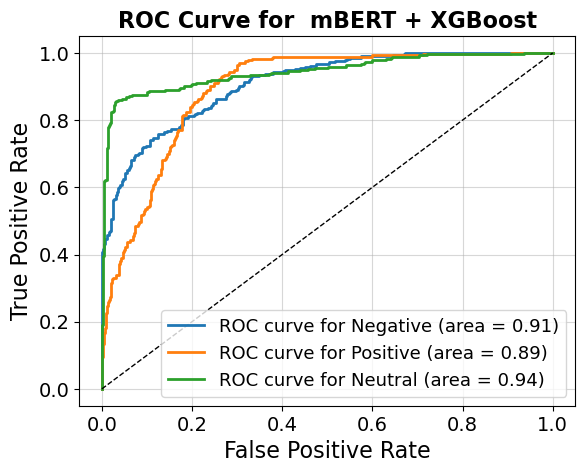}}
    \subfigure[]{\label{fig:g}\includegraphics[height=3cm, width=4.375cm]{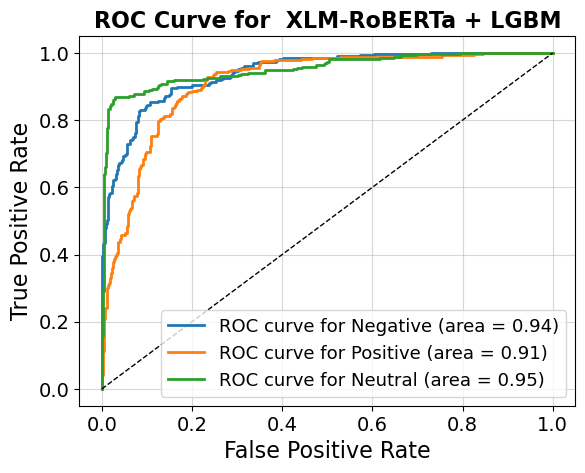}}
    \subfigure[]{\label{fig:h}\includegraphics[height=3cm, width=4.375cm]{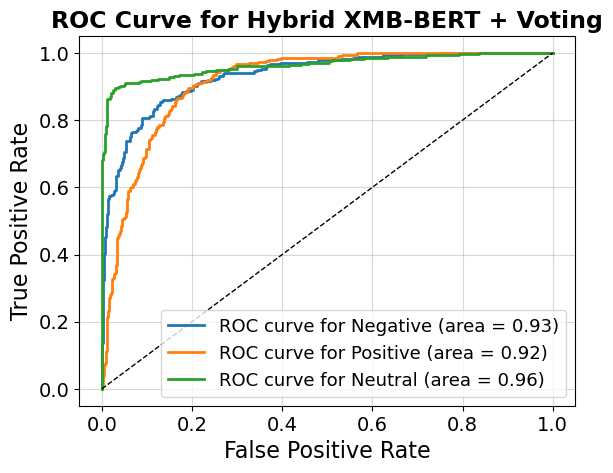}}
    \vspace{-0.5cm}
    \caption{Receiver-operating characteristic curve compare the performance of three different feature extraction techniques with Proposed Hybrid XMB-BERT: (a) BanglaBERT with Bagged SVM, (b) Multilingual BERT (mBERT) with Extreme Gradient Boosting (XGBoost, (c) XLM-RoBERTa with Light Gradient Boosting Machine (LGBM), and (d) Proposed Hybrid XMB-BERT with Voting Classifier, show the ROC Curve across three classes: 0, 1, 2, with values.} \vspace{-3mm}
    \label{fig:rc}
\end{figure*}

\subsection{Result Analysis}
\vspace{-2mm}The table \ref{tab:compare} compares the performance of the proposed Hybrid XMB-BERT model with BanglaBERT, mBERT, and XLM-RoBERTa across four evaluation metrics: Accuracy (A), Precision (P), Recall (R), and F1-score (F1). Each model was trained and tested using the same set of classifiers, ensuring a consistent and fair comparison. The results indicate that XMB-BERT outperforms the other models across all metrics and classifiers. Notably, with the Voting Classifier, XMB-BERT achieves the highest 83.7\% Accuracy, 84.1\% Precision, 83.7\% Recall, and 83.7\% F1-score, surpassing BanglaBERT 81.4\% accuracy with Bagged SVM, mBERT 78.4\% accuracy with XGBoost, and XLM-RoBERTa 82.2\% with LGBM in overall performance. Additionally, XMB-BERT consistently demonstrates superior performance with other classifiers, such as achieving 82.6\%, 82.5\%, 82.3\%, and 81.2\%  accuracy with XGB, LGBM, GB, and Bagged SVM, respectively, outperform BanglaBERT, mBERT, and XLM-RoBERTa with same classifiers. This evaluation highlights the effectiveness of the XMB-BERT architecture in sentiment classification by leveraging its ability to capture diverse linguistic features. \vspace{3mm}

\noindent Figure \ref{fig:cm} showcases the confusion matrices of four sentiment analysis models applied to Bangla social media comments on the July Revolution. Fig. \ref{fig:a} depicts the BanglaBERT with the Bagged SVM model, which correctly classifies 201 negative, 236 positives, and 246 neutral samples while displaying some confusion, misclassifying 62 negatives as positives and 41 positives as negatives. Fig. \ref{fig:b} presents mBERT with XGBoost, which slightly outperforms in classifying negative samples with 207 correct predictions but demonstrates higher misclassification rates for positive samples, including 42 negatives and 33 neutrals misclassified. Fig. \ref{fig:c} highlights XLM-RoBERTa with LGBM, which performs well overall by accurately classifying 232 negative, 223 positive, and 236 neutral samples, though it misclassifies 52 negatives as positives. Finally, Fig. \ref{fig:d} illustrates the proposed Hybrid XMB-BERT with Voting Classifier, which delivers the most balanced performance by correctly classifying 210 negative, 242 positive, and 246 neutral samples, with relatively fewer misclassifications, emphasizing its robustness across all sentiment categories. \vspace{2mm}

\noindent Figure \ref{fig:rc} illustrates the Receiver Operating Characteristic (ROC) curves for four combinations of feature extraction techniques and classifiers, highlighting the area under the curve (AUC) for three sentiment classes: Negative, Positive, and Neutral. Fig. \ref{fig:e} represents BanglaBERT + Bagged SVM, achieving an AUC of 0.92 for Negative, 0.90 for Positive, and 0.95 for Neutral, demonstrating strong discriminatory power across all classes. Fig. \ref{fig:f} shows mBERT + XGBoost, had slightly lower AUC values of 0.91 for Negative, 0.89 for Positive, and 0.94 for Neutral sentiments. Fig. \ref{fig:g} XLM-RoBERTa + LGBM achieved higher AUC values, with 0.94 for Negative, 0.91 for Positive, and 0.95 for Neutral sentiments. Finally, Fig. \ref{fig:h} displays the proposed XMB-BERT + Voting, which exhibited the best overall performance, with an AUC of 0.93 for Negative, 0.92 for Positive, and 0.96 for Neutral sentiments, indicating superior predictive capability. This comparison highlights that the XMB-BERT + Voting model consistently outperforms other combinations in distinguishing between sentiment classes, making it the most effective model in this evaluation.

\subsection{Discussion \& Limitations}
\vspace{-1mm} This research explores sentiment analysis of social media comments on the July Revolution using a Hybrid XMB-BERT model, comparing its performance with state-of-the-art transformer models, including BanglaBERT, mBERT, and XLM-RoBERTa. The results section demonstrates that XMB-BERT consistently outperforms other models across multiple evaluation metrics, showcasing its superior ability to capture diverse linguistic and contextual nuances in Bangla social media data. The ensemble-based voting classifier further boosts classification accuracy by harnessing the strengths of individual classifiers. \vspace{2mm}

\noindent Instead of depending on a single transformer model, our hybrid approach strategically combines both cross-lingual and language-specific embeddings. By integrating multiple transformer models, it enhances the system’s ability to capture richer semantic and syntactic features in Bangla, a low-resource language, particularly within the nuanced and emotionally charged context of political discourse. This architectural choice contributes to improved robustness and overall performance. Moreover, the ensemble learning mechanism balances classification accuracy with generalizability, allowing the model to adapt more effectively to the diverse linguistic patterns typical in Bangla social media content. \vspace{2mm}

\noindent Unlike well-studied domains such as English-language sentiment analysis or broader multilingual NLP tasks, there has been limited to no prior research focusing on Bangla-language sentiment analysis of political discourse on social media, particularly concerning the July Revolution. Moreover, our study introduces a completely new dataset and proposes a novel hybrid transformer-based model, which makes direct comparison with existing literature not feasible. The absence of benchmark datasets or similar methodological frameworks further underscores the gap in the field, highlighting the novelty and importance of our study in this underexplored domain. \vspace{2mm}

\noindent Despite these promising findings, the study is not without limitations. The dataset, while balanced and carefully labeled, was manually collected and exclusively focused on Bangla-language comments, which limits its generalizability to other domains or languages. Moreover, the model has not been tested across different social or political topics, making its adaptability and transferability to broader domains within Bangla NLP uncertain. Another limitation lies in the nature of the data itself: it does not include code-mixed or noisy social media text, which is common in real-world online discourse. As a result, the model’s robustness in less curated, more chaotic environments remains unverified. Additionally, although sentiment labels were carefully assigned, they are still susceptible to annotator bias and may not fully capture complex expressions such as sarcasm, irony, or ambiguous sentiment, which is especially prevalent in politically charged discussions. Finally, the hybrid model's reliance on multiple transformers increases computational complexity, requiring significant processing power and memory, factors that may constrain its scalability in resource-limited environments.

\section{CONCLUSION \& FUTURE WORK}
\vspace{-2mm} This research demonstrates the effectiveness of the proposed Hybrid XMB-BERT model for sentiment analysis of Bangla social media comments, especially in capturing subtle linguistic and contextual features during a sociopolitical movement. By integrating embeddings from mBERT, BanglaBERT, and XLM-RoBERTa, and optimizing feature representations through dimensionality reduction and ensemble classification, the proposed hybrid XMB-BERT achieves the highest accuracy of 83.7\%, along with superior precision, recall, and F1-score.  These findings not only contribute to the domain of Bangla-language sentiment analysis but also provide insights into the potential of hybrid transformer architectures for multilingual and context-specific text classification tasks. However, the model's limitations, such as computational complexity and domain specificity, point to several opportunities for future research. Future work will focus on expanding the dataset to include diverse domains and real-world social media variations such as code-mixed and noisy text, improving the model's adaptability and robustness. Additionally, incorporating multi-label classification for identifying hate speech, emotion, or sarcasm elements often intertwined with political discourse could further enhance interpretability and practical utility. To address computational challenges, lightweight transformer variants such as ALBERT and DistilBERT could be explored for efficiency without sacrificing accuracy. Moreover, integrating real-time data processing and temporal modeling would allow the system to track sentiment shifts dynamically, supporting use cases like public opinion monitoring during elections or social movements. Finally, embedding explainable AI (XAI) techniques into the framework would ensure transparency and foster user trust, especially among journalists, researchers, and policymakers seeking to understand model-driven insights. 

\section*{Data Availability}
\noindent The dataset used in this study, "Student-People Mass Uprising Public Sentiments Dataset", is publicly available for research purposes at \url{https://doi.org/10.5281/zenodo.15342899}. It includes 4,200 manually collected and labeled Bangla-language social media comments related to the 2024 July Revolution in Bangladesh. Researchers are encouraged to explore and use the dataset for further studies on sentiment analysis, Bangla NLP, and social media mining.

\section*{Responsibility and Ethical Considerations}
\noindent We acknowledge the potential risks and limitations associated with using social media data for research purposes, particularly in politically sensitive contexts. While the dataset used in this study was carefully curated and is now publicly available to ensure transparency and reproducibility, we recognize that users' content may introduce noise, misinformation, or bias. Although a formal fact-checking step was not included due to the subjective and informal nature of social media comments, we mitigated these concerns through careful manual annotation. The authors take full responsibility for any potential issues or misinterpretations arising from this dataset and strongly encourage future researchers to apply additional verification or filtering methods tailored to their specific use cases.


\bibliographystyle{ieeetr}
\bibliography{ref.bib}

\end{document}